\documentclass[a4paper, 10 pt, conference]{ieeeconf}      
\IEEEoverridecommandlockouts                              
\usepackage{graphicx}
\usepackage{amssymb}
\usepackage{amsmath}
\usepackage{array}
\usepackage{color}
\usepackage{bm}
\usepackage{times}
\usepackage[table]{xcolor}
\usepackage{optidef}
\usepackage{multirow}
\usepackage{enumitem}
\usepackage{booktabs}
\usepackage{url}
\usepackage{setspace}
\usepackage[noadjust]{cite}
\usepackage[linesnumbered,ruled,vlined]{algorithm2e}

\SetAlFnt{\small}
\SetAlCapFnt{\normalsize}
\SetAlCapNameFnt{\normalsize}
\usepackage{algorithmic}
\algsetup{linenosize=\tiny}

\xdefinecolor{gray95}{gray}{0.65}
\xdefinecolor{gray92}{gray}{0.9}
\xdefinecolor{gray25}{gray}{0.8}
\newcolumntype{L}[1]{>{\raggedright\let\newline\\\arraybackslash\hspace{0pt}}m{#1}}
\newcolumntype{C}[1]{>{\centering\let\newline\\\arraybackslash\hspace{0pt}}m{#1}}
\newcolumntype{R}[1]{>{\raggedleft\let\newline\\\arraybackslash\hspace{0pt}}m{#1}}

\title{\LARGE \bf
Deep Echo State Networks for Short-Term Traffic Forecasting: \\Performance Comparison and Statistical Assessment}

\author{Javier Del Ser$^{1,2}$, Ibai La\~ na$^{1}$, Eric L. Manibardo$^{1}$, Izaskun Oregi$^{1}$, Eneko Osaba$^{1}$, \\Jesus L. Lobo$^{1}$, Miren Nekane Bilbao$^2$, Eleni I. Vlahogianni$^{3}$
\thanks{$^{1}:$ TECNALIA, Basque Research and Technology Alliance (BRTA), 48160 Derio, Spain. Contact email: javier.delser@tecnalia.com}%
\thanks{$^{2}:$ University of the Basque Country (UPV/EHU), 48013 Bilbao, Spain.}%
\thanks{$^{3}:$ National Technical University of Athens (NTUA), 157 73 Greece.}%
}

\begin{document}
\bstctlcite{IEEEexample:BSTcontrol}
\maketitle
\thispagestyle{empty}
\pagestyle{empty}

\begin{abstract}
In short-term traffic forecasting, the goal is to accurately predict future values of a traffic parameter of interest occurring shortly after the prediction is queried. The activity reported in this long-standing research field has been lately dominated by different Deep Learning approaches, yielding overly complex forecasting models that in general achieve accuracy gains of questionable practical utility. In this work we elaborate on the performance of Deep Echo State Networks for this particular task. The efficient learning algorithm and simpler parametric configuration of these alternative modeling approaches make them emerge as a competitive traffic forecasting method for real ITS applications deployed in devices and systems with stringently limited computational resources. An extensive comparison benchmark is designed with real traffic data captured over the city of Madrid (Spain), amounting to more than 130 Automatic Traffic Readers (ATRs) and several shallow learning, ensembles and Deep Learning models. Results from this comparison benchmark and the analysis of the statistical significance of the reported performance gaps are decisive: Deep Echo State Networks achieve more accurate traffic forecasts than the rest of considered modeling counterparts.
\end{abstract}

\section{Introduction}

Much has been said about traffic forecasting over the years, with particular profusion in the last decade as evinced by recent prospective overviews on this matter \cite{lana2018road}. The unprecedented scales at which data is currently captured from the road network has spawned a myriad of intelligent applications exploiting such information flows, from traffic condition monitoring to intelligent personal routing assistants, among others \cite{zhang2011data}. Spurred by this rich ecosystem of data-based traffic services, the research community has devoted huge efforts towards delivering predictive insights on how traffic flows evolve over time, including its impact on endogenous and exogenous aspects to traffic itself, from congestion prediction \cite{jabbarpour2018applications} to environmental pollution assessment \cite{lana2016role}, to mention a few.

In this context, the delivery of such predictive insights from traffic data has progressively shifted from traditional methods for time series forecasting \cite{karlaftis2011statistical}, to more elaborated modeling approaches relying on concepts from supervised Machine Learning (ML, \cite{angarita2019taxonomy}). As such, methods falling within this family of modeling approaches aim at capturing correlation patterns between a set of inputs and a target variable to be predicted based on a set of annotated data instances (\emph{examples}). Once this pattern has been learned, the trained model at hand is assumed to generalize well to unseen examples, and is thereafter utilized for predicting future values of the target traffic variable. Beyond their use as a replacement of traditional forecasting techniques, supervised learning has been also exploited to infer complex spatio-temporal relationships from traffic data \cite{min2011real,ermagun2018spatiotemporal}. Indeed, the extension of supervised modeling to multivariate time series forecasting can be performed in an straightforward fashion, by simply adding new features to the model input. By contrast, the incorporation of spatial information to the predictive model has been recently shown to be approachable by means of graph embedding techniques, allowing spatial information of the road network to be incorporated into the predictive model at hand \cite{peng2020spatial,xie2019sequential}.

When placing the learning algorithms underneath supervised ML under the spotlight, a quick inspection of the recent state of the art reveals that traffic forecasting has lately become almost capitalized by different flavors of Deep Neural Networks (DNN). The new forms of neural computation behind this paradigm have certainly given rise to unprecedented levels of predictive performance, particularly in sequence regression \cite{shi2019survey,polson2017deep}. As a result, other algorithmic alternatives have grasped relatively less attention -- or have even been neglected -- by the community working on traffic forecasting. For instance, even though time series are particularly suited for their representation with temporal spike trains, the research activity noted around the use of Spiking Neural Networks (SNN) has been so far evidenced by a scarcity of contributions \cite{lana2019adaptive,lana2018roadsnn}. 

In this paper we pause at one of the algorithmic families that has arguably remained overlooked for traffic forecasting until very recently: Reservoir Computing (RC) \cite{schrauwen2007overview}. RC denotes a particular branch of randomization-based neural networks composed by a set of sparsely connected neural processing units (\emph{reservoir}), which are linked to the target variable of interest through weighted (learnable) connections and a readout layer. This is the design principle shared by popular RC models such as Echo State Networks (ESNs, \cite{jaeger2001echo}) and Liquid State Machines \cite{maass2002real}, which have been put to practice in dozens of forecasting problems in energy grids \cite{hamedani2018reservoir}, prognosis \cite{long2019evolving,li2019adaptive} or medicine \cite{buteneers2011automatic}. Most of the literature adopting RC models agree on the rationale for selecting them over Deep Learning approaches: 1) their computationally mode affordable training algorithm, overriding any need for specialized computational resources (such as GPU); and 2) their simpler structure, which simplifies the initially model handcrafting process from which DNN modeling departs, and avoids any need for resorting to automatic design methods \cite{martin2018evodeep,suganuma2017genetic}. 

In this work we take a step further over previously reported works dealing with the application of RC methods to traffic forecasting. To the best of our knowledge, the first contribution in this direction was made in \cite{an2011short}, where a naive ESN model was used to forecast traffic over a single urban road section in Xian (China). A more complete study comprising stacking ensembles of ESNs was published recently in \cite{del2019road}, including a comparison to other supervised learning methods. Results obtained in this latter work lead to a conclusion of utmost interest for the purpose of the present study: no statistically significant performance gaps were found between stacking ESN ensembles and recurrent Long Short-Term Memory (LSTM) networks, concluding, with empirical evidences, that RC is a computationally efficient modeling choice for traffic forecasting. 

This paper continues along this research path by exploring the suitability of brand new multilayered forms of RC (specifically, Deep ESN \cite{gallicchio2017deep,gallicchio2018design}) for short-term traffic forecasting. We advocate for this research direction in light of reflective studies \cite{gallicchio2018layering}, which postulate the suitability of these RC models over recurrent DNN architectures. To this end, we design an extensive experimental benchmark over real traffic volume data captured by more that 130 inductive sensors deployed over the city of Madrid (Spain). Unlike other studies reported in recent times, we further implement a principled comparison methodology that allows verifying the statistical significance of existing gaps among the traffic forecasting models under comparison. Results are conclusive, paving the way towards the incorporation of Deep ESN models in applications with traffic forecasting at their core.

The remainder of this manuscript is structured as follows: for the sake of self-completeness, first Section \ref{sec:reservoir} provides a brief introduction to the fundamentals of ESNs, whereas the details underneath Deep ESN are given in Section \ref{sec:proposed}, along with an explanation of its particular training procedure. Next, Section \ref{sec:exp_setup} presents the experimental setup and comparison methodology. Section \ref{sec:results} discusses the obtained performance results and their statistical analysis. Section \ref{sec:conc} draws conclusions and future research paths rooted on this work.

\section{Fundamentals of Echo State Networks} \label{sec:reservoir}

In general, RC models build upon the capability of Recursive Neural Networks (RNN) to perform well even when their constituent trainable parameters (weights) are not tailored for the predictive modeling task under consideration \cite{schrauwen2007overview}. Interestingly, a set of randomly connected recurrent neurons suffice for mapping the input data to a space of recurrent states, composing what is known as the \emph{reservoir} that mints the name of this family of learning algorithms. The states of the reservoir are then mapped to the output of the model via a \emph{readout layer}, which usually is set to a linear regression model with a regularization mechanism.

\begin{figure}[h!]
	\vspace{-1mm}
	\centering
	\includegraphics[width=0.87\columnwidth]{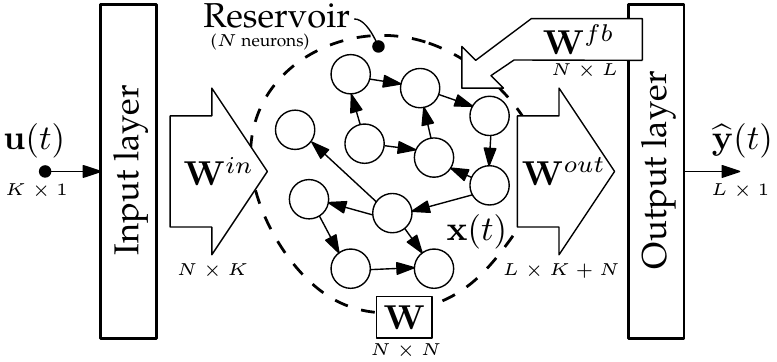}
	\vspace{-1mm}
	\caption{Architecture of a generic ESN model.}
	\label{fig:reservoir_computing}
	\vspace{-1mm}
\end{figure}

An ESN is one particular RC modeling flavor that learns to map a $K$-dimensional input $\mathbf{u}(t)$ to a $L$-dimensional output $\mathbf{y}(t)$ throughout a reservoir of $N$ units. As shown in Figure \ref{fig:reservoir_computing}, the reservoir states at time $t$ are recurrently updated as:
\begin{align*} \label{eq:esn_1}
\mathbf{x}(t+1) = \alpha f(\mathbf{W} \mathbf{x}(t) + \mathbf{W}^{in} \mathbf{u}(t+1) + \mathbf{W}^{fb} \mathbf{y}(t)) + (1-\alpha)\mathbf{x}(t),
\end{align*}
where $t=1,\ldots, T$ stands for a discrete time index, $\mathbf{x}(t)$ denotes the $N$-dimensional state vector of the reservoir; $f(\cdot)$ is an activation function; \smash{$\mathbf{W}_{N\times N}$, $\mathbf{W}^{in}_{N\times K}$ and $\mathbf{W}^{fb}_{N\times L}$} are weight matrices; and $\alpha\in\mathbb{R}(0,1]$ is the so-called leaking rate, which imprints different update dynamics in $\mathbf{x}(t)$. 

Once this recurrence is applied at time $t$, the output of the ESN model is given by:
\begin{equation} \label{eq:esn_2}
\widehat{\mathbf{y}}(t) = g\left(\mathbf{W}^{out} [\mathbf{x}(t);\;\mathbf{u}(t)]\right),
\end{equation}
where $[\cdot;\cdot]$ denotes vector concatenation, $\mathbf{W}^{out}_{L\times(K+N)}$ is a \emph{readout} matrix containing the output weights, and $g(\cdot)$ is an activation function. The weights of the matrices involved in the computation can be iteratively adjusted as per a training dataset with example(s) $\{(\mathbf{u}(t),\mathbf{y}(t))\}$. Analogously to their DNN counterparts, this iterative adjustment can be done by minimizing a loss function $L(\mathbf{y};\widehat{\mathbf{y}})$ that quantifies the error between the output $\widehat{\mathbf{y}}(t)$ of the ESN model, and its corresponding target output $\mathbf{y}(t)$. For regression tasks, the root mean square error (RMSE) loss is typically used:
\begin{equation}
L(\mathbf{y};\widehat{\mathbf{y}})=\frac{1}{L} \sum_{l=1}^L \left(\frac{1}{T}\sum_{t=1}^T(y_l(t)-\widehat{y}_l(t))^2\right)^{1/2},
\end{equation}
where the RMSE error is averaged first over time and second, over the dimensions $l\in\{1,\ldots,L\}$ of the ESN output $\mathbf{y}(t)=[y_1(t),\ldots,y_l(t),\ldots,y_L(t)]$. Although there are clear similarities between this weight adjustment process and the gradient backpropagation performed in DNN models for the same purpose, a crucial difference exists between how ESN models are fitted in practice with respect to other forms of neural computation: not all weight matrices $\mathbf{W}^{in}$, $\mathbf{W}$, $\mathbf{W}^{fb}$ and $\mathbf{W}^{out}$ are adjusted. Instead, the weight values of input, hidden state, and feedback matrices are drawn initially at random, leaving the weights of the output matrix $\mathbf{W}^{out}$ as the only ones that are adjusted during the training phase. To this end, a least-squares linear regression can be employed:
\begin{equation}\label{eq:Wout}
\min\limits_{\mathbf{w}^{out}_{l}\in\mathbb{R}^{K+N}} \sum_{t=1}^T \left(\sum_{j=1}^{K+N}w^{out}_{l,j}\cdot [\mathbf{x}(t);\;\mathbf{u}(t)]_j-y_l(t)\right)^2,
\end{equation}
where $l\in\{1,\ldots,L\}$, $[\cdot;\cdot]_j$ returns the $j$-th element of the concatenation, and $\mathbf{w}^{out}_{l}=[w^{out}_{l,1},\ldots,w^{out}_{l,K+N}]^T$ is the $l$-th row of $\mathbf{W}^{out}$. In order to prevent the model from overfitting, regularized least squares versions are often enforced, such as Ridge Regression with Tikhonov-Phillips regularization:
\begin{equation}\label{eq:Wout_reg}
\min\limits_{\mathbf{w}^{out}_{l}\in\mathbb{R}^{K+N}} \sum_{t=1}^T \left(\sum_{j=1}^{K+N}w^{out}_{l,j}\cdot z_j(t)-y_l(t)\right)^2\hspace{-2mm}+\lambda ||\mathbf{w}^{out}_{l}||_2^2,
\end{equation}
where $||\cdot||_2$ denotes $L_2$ norm, and parameter $\lambda\in\mathbb{R}[0,\infty)$ establishes the relative importance of the $L_2$ regularization term in the minimization problem. Therefore, neurons inside the repository can be conceived as randomly chosen echoes modeling different temporal patterns over the input signal $\mathbf{x}(t)$, whereas the readout layer as per \eqref{eq:Wout} serves as a mapping between such echoes and the target signal $\mathbf{y}(t)$. 

\section{Deep Echo State Networks} \label{sec:proposed}

Although off-the-sheld ESN models have been shown to render competitive performance scores in a manifold of regression problems, a great deal of attention has been lately paid to more sophisticated multi-layered RC models. Indeed, several works have advanced the potentiality of hierarchically organized temporal features produced by stacked recurrent neural networks to represent patterns at different time scales \cite{1312.6026}. Deep ESN capitalizes this concept by stacking a number of ESN models, such that each model is fed by the output of the preceding one in the stack. As a result of this layered stacking of reservoirs, the global state vector resulting from the concatenation of the different state vectors $[\mathbf{x}^{(n)}(t)]_{n=1}^{N_L}$ represents the history of the input $\mathbf{u}(t)$ at multiple time scales, even if reservoirs located inside each layer are configured similarly in terms of their hyper-parameters. This multi-scale property, along with the increased richness and diversity of reservoir dynamics attained by stacked ESNs \cite{gallicchio2019richness}, gives rise to a modeling approach that may contend competitively with DNNs.

Following Figure \ref{fig:deep_ESN} and the notation introduced above, superindex $\scriptsize{(n)}$ hereafter denotes that the parameter at hand belongs to the $n$-th layer of the Deep ESN model. Therefore, the update policy of the first stacked ESN is given by:
\begin{align} \label{eq:deepesn_1}
&\mathbf{x}^{(1)}(t+1) = \left(1-\alpha^{(1)}\right)\mathbf{x}^{(1)}(t) + \nonumber \\
&\alpha^{(1)} f\left(\mathbf{W}^{(1)} \mathbf{x}^{(1)}(t) + \mathbf{W}^{(1),in} \mathbf{u}^{(1)}(t+1)\right),
\end{align}
where dimensions of weight matrices are kept to the same than in the single ESN case. For layers $1<n\leq N_L$, the updating recurrence can be generalized as:
\begin{align} \label{eq:deepesn_2}
&\mathbf{x}^{(n)}(t+1) = \left(1-\alpha^{(n)}\right)\mathbf{x}^{(n)}(t) + \nonumber \\
&\alpha^{(n)} f\left(\mathbf{W}^{(n)} \mathbf{x}^{(n)}(t) + \mathbf{W}^{(n),in} \mathbf{x}^{(n-1)}(t+1)\right),
\end{align}
from where the output of the overall Deep ESN model is computed as:
\begin{equation} \label{eq:deepsn_3}
\widehat{\mathbf{y}}(t) = g\left(\mathbf{W}^{out} [\mathbf{x}^{(1)}(t),\ldots,\mathbf{x}^{(N_L)}(t)]\right),
\end{equation}
where $\mathbf{W}_{L\times (N_L \cdot N)}^{out}$ is the readout weight matrix that maps the internal states of the entire set of stacked reservoirs to the output signal of interest. Unlike in the output layer of the single ESN model as per Expression \eqref{eq:esn_2}, no copy of the input $\mathbf{u}(t)$ is considered in the concatenated state vector.
\begin{figure}[h!]
	\vspace{-1mm}
	\centering
	\includegraphics[width=0.82\columnwidth]{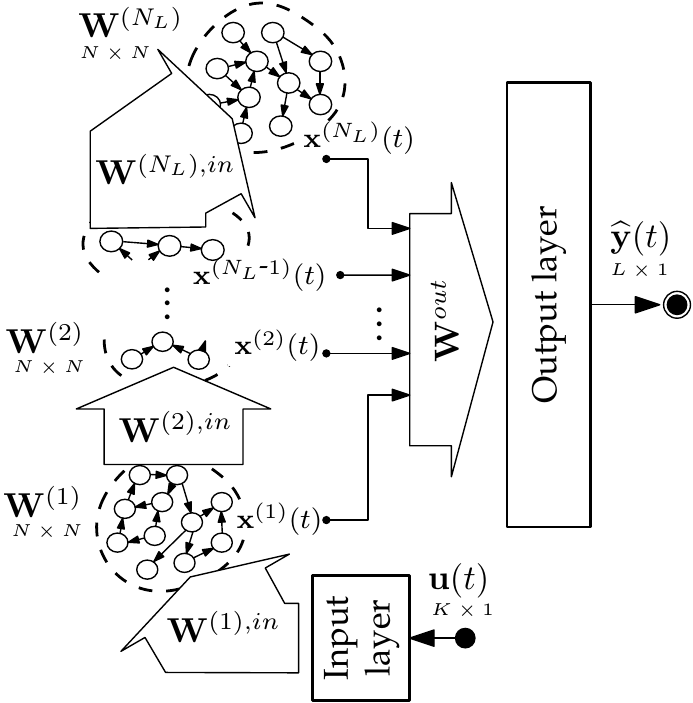}
	\vspace{-3mm}
	\caption{Stacked multilayered architecture of a Deep ESN model.}
	\label{fig:deep_ESN}
	\vspace{-1mm}
\end{figure}

Similarly to single ESN models, reservoir weights represented by the weight matrices $\mathbf{W}^{(n)}$ and $\mathbf{W}^{(n),in}$ are initialized at random for each layer $n=1,\ldots,N_L$ (e.g. by uniform sampling over $\mathbb{R}[-1,1]$) and rescaled to fulfill the so-called Echo State Property, i.e.:
\begin{equation} \label{eq:deepsn_esp}
\max _{n=1,\ldots,N_{L}} \rho\left((1-\alpha^{(n)}) \mathbf{I}+\alpha^{(n)} \mathbf{W}^{(n)}\right)<1,
\end{equation}
with $\rho$ denoting the largest absolute eigenvalue of its argument matrix. Once weight values for $\mathbf{W}^{(n)}$ and $\mathbf{W}^{(n),in}$ have been set, they are left untrained for the rest of the training process. The weight matrix $\mathbf{W}^{out}$ of the output layer is adjusted on the training set at hand by means of regularized least-squares methods as the one in \eqref{eq:Wout_reg}, or other direct solving technique such as Moore-Penrose pseudo-inversion.

\section{Experimental Setup} \label{sec:exp_setup}

\begin{figure}[h!]
	\centering
	\vspace{-2mm}
	\includegraphics[width=0.75\columnwidth]{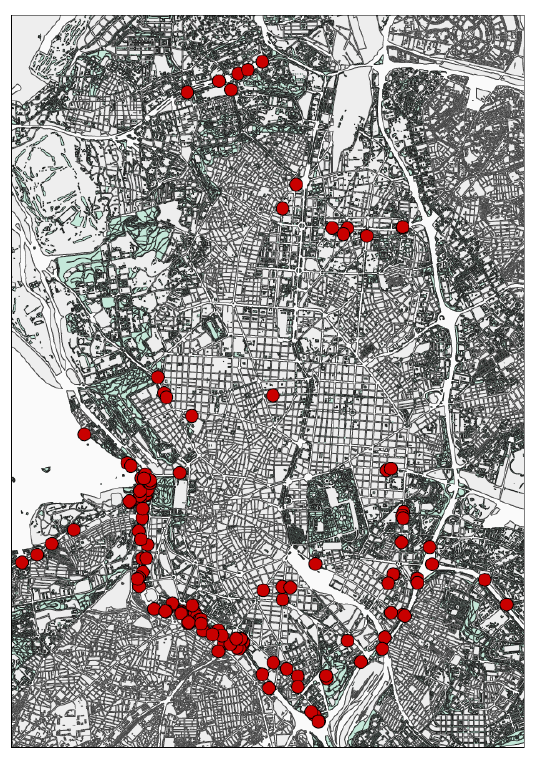}
	\caption{Location of selected ATRs in the city of Madrid (Spain).}
	\vspace{-1mm}
	\label{Fig:ubicacion}
\end{figure}

In order to comparatively assess the performance of Deep ESN when applied to short-term traffic forecasting, we devise an experimental setup involving real data collected during 2017 in streets, urban arterials and freeways of the city of Madrid (Spain). A public data source maintained by the City Council of Madrid\footnote{\url{http://datos.madrid.es}. Accessed on February 22nd, 2020.} has been used, providing traffic flow readings of around 4,000 Automatic Traffic Readers (ATRs) deployed over the city road network. Data are presented in historic datasets with traffic flow readings aggregated every 15 minutes, yielding a maximum of 35,040 observations per year and location. In order to avoid problems related to missing data (which is out of the scope of this research), we have selected a number of locations that present less than 3\% of missing data during the surveyed period. Thus, simple imputation mechanisms are enough to make the dataset complete \cite{lana2018imputation}. This yields a selection of $N_{ATR}=133$ ATRs which, as displayed in Figure \ref{Fig:ubicacion}, are distributed in diverse areas, covering from freeways to residential or business areas, featuring very different traffic profiles. Moreover, each model under comparison is evaluated for 4 different prediction horizons $h\in\{1,2,3,4\}$ [slots], so as to gauge their performance as the target variable is farther in time. 

The benchmark is composed by an assorted mixture of $M=13$ supervised learning models for traffic forecasting, composed by 1) traditional regression methods (least-squares linear regression [\texttt{LR}], k Nearest Neighbors [\texttt{kNN}], Decision Tree [\texttt{DTR}], Extreme Learning Machine [\texttt{ETR}], and $\varepsilon$-Support Vector Machine [\texttt{SVM}]); 2) ensembles (Adaboost [\texttt{ADA}], Random Forest [\texttt{RFR}], Extremely Randomized Trees [\texttt{ETR}] and Gradient Boosting [\texttt{GBR}]); and 3) neural networks (a Multi-Layer Perceptron [\texttt{MLP}], a recurrent neural network based on LSTM units [\texttt{LSTM}], and a Deep ESN model [\texttt{DeepESN}]). This comparison benchmark is complemented by a naive persistence approach which assigns the last known value of the target variable at time $t$ as the prediction for time $t+h$. The consideration of this latter approach permits to evaluate the real predictive gains achieved by more elaborated modeling methods. 

Hyper-parameters of all the aforementioned models (including the topology design of the neural networks under consideration) have been tailored beforehand by means of an exhaustive search over a fine-grained grid of values. Each hyper-parameter combination was evaluated over a separate subset of $10$ ATRs in terms of their average coefficient of determination ($R^2$). The average score was computed over a $10$-fold time-split cross-validation of the time series of these subset of ATRs, which was reframed as a supervised learning problem by a sliding window of $W=6$ past samples. Since the discussion is held on results aggregated over ATRs with varying traffic profiles, in what follows we resort this relative score rather than other absolute performance metrics. 
\vspace{-2mm}
\begin{algorithm}[h!]
	\DontPrintSemicolon
	\SetKwIF{If}{ElseIf}{Else}{If}{}{ElseIf}{Else}{end}
	\SetKwFor{For}{For}{}{end}
	\KwIn{Traffic forecasting models $\{M_{\varphi_m}^m\}_{m=1}^M$ with hyper-parameters $\vartheta_m$, forecasting horizon $h$, traffic data \smash{$\{\mathbf{u}_a(t)\}_{a=1}^{N_{ATR}}$}, significance level $\alpha$}
	\KwOut{Average ranking $\{avgRank_m\}_m^M\in\mathbb{R}[1,K]$ and critical distance $CD$}
	\For{$a=1,\ldots,N_{ATR}$}
	{
		Transform $\mathbf{u}_a(t)$ to supervised learning by using a $W$-sized sliding window of recent values\;
		Compute $P$ train/test time splits of $\mathbf{u}_a(t)$\;
		\For{$m=1,\ldots,M$}{
			\For{$p=1,\ldots,P$}{
				Train and test model $M_{\vartheta_m}^m$ on partition $k$\;
				Compute performance metric $score_{m}^{p,a}$\;
			}
			Compute $avgScore_{m}^a=1/K\sum_{p=1}^P score_{m}^{p,a}$\;
			Set $\texttt{WINS}_{m}^a=\texttt{TIES}_{m}^a=\texttt{LOSSES}_{m}^a=0$
		}
		\If{$\texttt{Friedman}(\{\text{avgScore}_{m}^a\}_{m=1}^M,\alpha)$}
		{ 
			Set $rank_{m}^{a}$ to $(1/M)\sum_{m=1}^M m$ $\forall m$\;
		}
		\Else{
			\For{$m,m' \in \{1,\ldots,M\}\times \{1,\ldots,M\}:\hspace{1mm} m<m'$}
			{
				\If{$\texttt{Wilcoxon}(\{score_{m}^{p,a},score_{m'}^{p,a}\}_{p=1}^P,\alpha)$}
				{
					$\texttt{TIES}_{m}^a+=1$, $\texttt{TIES}_{m'}^a+=1$\;
				}
				\ElseIf{$avgScore_{m}^a>avgScore_{m'}^a$}
				{
					$\texttt{WINS}_{m}^a+=1$, $\texttt{LOSSES}_{m'}^a+=1$\;
				}
				\Else
				{
					$\texttt{WINS}_{m'}^a+=1$, $\texttt{LOSSES}_{m}^a+=1$\;
				}
			}
			Compute fractional rankings $\{rank_{m}^{a}\}_{m=1}^M$ as per $\{\texttt{WINS}_{m}^a,\texttt{TIES}_{m}^a\}_{m=1}^M$\;
		}
	}
	\For{$m=1,\ldots,M$}{
		$avgRank_m = (1/N_{ATR})\sum_{a=1}^{N_{ATR}} rank_{m}^{a}$\;
	}
	Compute $CD$ as per Expression \eqref{eq:cd}\;
	\caption{Comparison methodology}
	\label{alg:comparison}
\end{algorithm}
\begin{table*}[h!]
	\centering
	\caption{Mean$\pm$standard deviation of the $R^2$ score for each traffic forecasting model under consideration} \label{tbl:scores}
	\resizebox{2\columnwidth}{!}{
		\begin{tabular}{ccccccclcccclccc}
			\toprule
			& & \multicolumn{5}{c}{Shallow learning methods} & & \multicolumn{4}{c}{Ensembles} & & \multicolumn{3}{c}{Neural networks} \\
			\cmidrule{3-7} 	\cmidrule{9-12} \cmidrule{14-16}
			Horizon & Persistence & \texttt{LR} & \texttt{kNN} & \texttt{DTR} & \texttt{ELR} & \texttt{SVR} & & \texttt{ADA} & \texttt{RFR} & \texttt{ETR} & \texttt{GBR} & & \texttt{MLP} & \texttt{LSTM} & \texttt{DeepESN} \\
			\midrule
			$h=1$ & 0.83$\pm$0.14 & 0.79$\pm$0.12 & 0.81$\pm$0.13 & 0.81$\pm$0.13 & 0.81$\pm$0.13 & 0.83$\pm$0.11 & & 0.75$\pm$0.13 & 0.82$\pm$0.12 & 0.82$\pm$0.12 & 0.83$\pm$0.13 & & 0.83$\pm$0.13 & 0.83$\pm$0.12 & 0.87$\pm$0.11 \\
			$h=2$ & 0.76$\pm$0.14 & 0.73$\pm$0.12 & 0.76$\pm$0.14 & 0.77$\pm$0.12 & 0.77$\pm$0.12 & 0.79$\pm$0.12 & & 0.70$\pm$0.13 & 0.78$\pm$0.13 & 0.78$\pm$0.13 & 0.79$\pm$0.13 & & 0.79$\pm$0.13 & 0.79$\pm$0.13 & 0.84$\pm$0.12\\
			$h=3$ & 0.69$\pm$0.14 & 0.66$\pm$0.13 & 0.72$\pm$0.14 & 0.73$\pm$0.13 & 0.71$\pm$0.13 & 0.74$\pm$0.12 & & 0.66$\pm$0.13 & 0.74$\pm$0.14 & 0.74$\pm$0.14 & 0.75$\pm$0.14 & & 0.75$\pm$0.13 & 0.77$\pm$0.13 & 0.81$\pm$0.11 \\
			$h=4$ & 0.61$\pm$0.14 & 0.59$\pm$0.13 & 0.67$\pm$0.15 & 0.67$\pm$0.14 & 0.66$\pm$0.13 & 0.68$\pm$0.13 & & 0.62$\pm$0.12 & 0.69$\pm$0.14 & 0.69$\pm$0.14 & 0.70$\pm$0.14 & & 0.70$\pm$0.13 & 0.69$\pm$0.13 & 0.77$\pm$0.15 \\
			\bottomrule
	\end{tabular}}
\vspace{-4mm}
\end{table*}

A procedure is needed to rank the different models for every prediction horizon $h\in\{1,2,3,4\}$. For this purpose, a solid criterion must be first designed to decide, with statistical significance, whether a model outperforms another given their performance scores computed for every ATR. Algorithm \ref{alg:comparison} summarizes this methodology. Given a prediction horizon $h$ and an ATR, $R^2$ scores were recorded for every model over $10$ train-test time-split partitions, providing a more confident estimation of the performance of the model than other cross-validation strategies \cite{sheridan2013time}. Once these cross-validated scores were produced, two statistical hypothesis tests were applied at a significance level $\alpha=0.05$: 
\begin{enumerate}[leftmargin=*]
\item A Friedman test \cite{pereira2015overview}, to enforce an initial check whether statistically significant gaps are present between the average cross-validated scores obtained for the models. 
\item A Wilcoxon signed rank test \cite{wilcoxon1992individual}, to assess the relevance of the difference among means in pairwise comparisons among different models. By iterating over every pair of models, counters recording the number of \texttt{WINS}, \texttt{TIES} and \texttt{LOSSES} of every modeling choice can be obtained, so that models can be ranked in descending order of their \texttt{WIN} counter. Ties are resolved by fractional ranking, so that tied learners are assigned the same ranking number equal to the mean of what should be their ordinal rankings. It is important to emphasize that both \texttt{TIES} and  \texttt{WINS}/\texttt{LOSSES} are decided with statistical significance, as dictated by the Wilcoxon signed rank test.
\end{enumerate}

Finally, once ranks are computed for the 133 ATRs, we use the average ranking of each model to compute the so-called \emph{Nemenyi critical distance} \cite{nemenyi} with $\alpha=0.05$. The critical distance imposes the minimum difference among average ranks of the models under comparison for their differences to be significant, and can be computed as:
\begin{equation}\label{eq:cd}
CD = Q_{\alpha,M} \sqrt{\frac{M(M+1)}{6 N_{ATR}}},
\end{equation}
where $M$ is the number of models, $N_{ATR}$ the number of datasets, and critical values $Q_{\alpha,M}$ can be computed from tabulated values of the Studentized range statistic for infinite degrees of freedom divided by $\sqrt{2}$.

\section{Results, Discussion and Statistical Analysis} \label{sec:results}

We begin our discussion by pausing at Table \ref{tbl:scores}, which summarizes the mean and standard deviation of the $R^2$ scores attained by the models under comparison over the $133$ ATRs. 

First it is important to note that \texttt{DeepESN} consistently dominates the benchmark, achieving average $R^2$ scores notably superior than the rest of models. However, it is relevant to highlight that for low values of the prediction horizon, the naive persistence approach is able to attain comparable performance scores to those achieved by other models, which degrades as the prediction horizon increases. This result evidences the existence of a strong amount of autocorrelation in the traffic time series of the ATRs, which outweighs the use of most supervised learning models considered in the benchmark. \texttt{DeepESN}, however, maintains a superior average performance even for short prediction horizons, by virtue of its superior capability to represent long term relationships. Interestingly, the \texttt{LSTM} model also seems to leverage its trainable memory to capture long-term temporal patterns from data, rendering higher average $R^2$ values than the persistence approach, yet surpassed by other methods in the benchmark (specially \texttt{MLP} and \texttt{GBR}) and significantly outperformed by \texttt{DeepESN}. However, the relatively high value of the standard deviation, and the fidelity of its estimation computed over 133 ATRs, calls for the aforementioned analysis to elicit guarantees of the statistical relevance of these identified performance gaps.

\begin{figure}[h!]
	\centering
	\vspace{-2mm}
	\includegraphics[width=0.92\columnwidth]{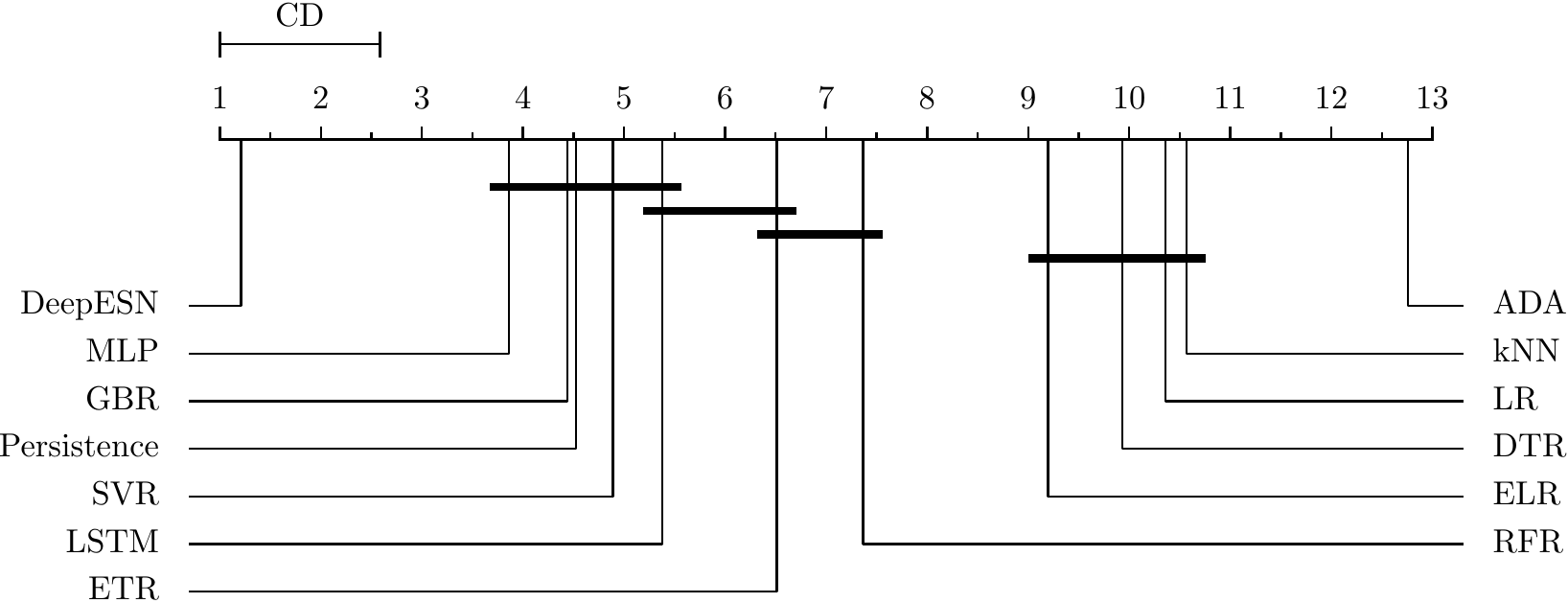}\\
	(a) $h=1$ \\
	\includegraphics[width=\columnwidth]{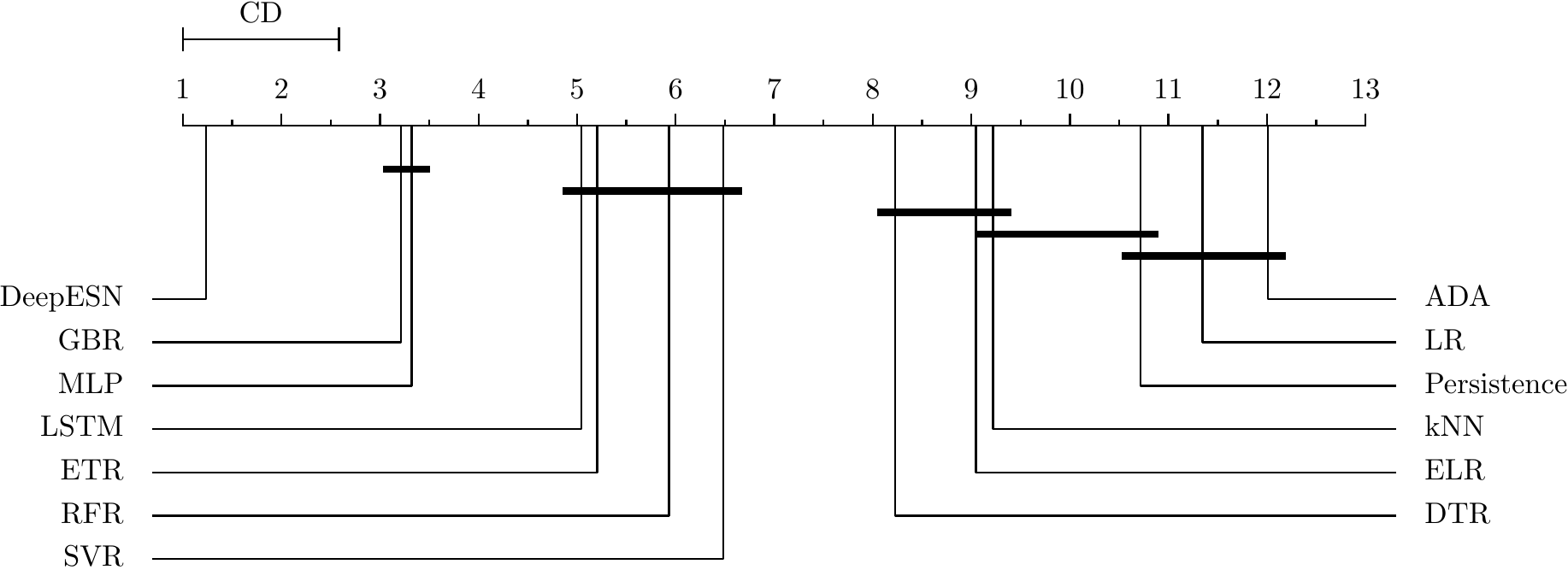}\\
	(b) $h=4$ 
	\vspace{-1mm}
	\caption{Critical distance plot of the average rankings for (a) $h=1$; (b) $h=4$. CD stands for critical distance. The cases $h=2$ and $h=3$ lead to identical conclusions and are not shown due to space constraints.}
	\label{fig:cddiagrams}
\end{figure}

The results of this statistical analysis are summarized graphically as \emph{critical distance plots} in Figures \ref{fig:cddiagrams}.a ($h=1$) and \ref{fig:cddiagrams}.b ($h=4$). These plots depict the average ranking of the models in the benchmark output by the methodology in Algorithm \ref{alg:comparison}, along with the critical distance $CD$ that determines the minimum average ranking gap among models for their performance gaps to be considered statistically significant \cite{demvsar2006statistical}. Therefore, algorithms joined by a bold horizontal line must be regarded as equivalent to each other. As elucidated by these plots, the outperforming behavior of \texttt{DeepESN} identified in Table \ref{tbl:scores} results to be statistically significant, whereas no clear \emph{second best} can be decided as per the power of the statistical tests in use. 

These insightful results must be jointly appraised with the relatively low parametric space and high computational efficiency of \texttt{DeepESN}, which must be regarded as additional advantages that contribute to the suitability of this modeling choice for short-term traffic forecasting.

\section{Conclusions and Future Directions} \label{sec:conc}

In this manuscript we have explored the performance of Deep ESN models for short-term traffic forecasting. Our research hypothesis departs from the renowned capability of multilayered structures of recurrent processing units to represent patterns at multiple time scales, which arguably matches the modeling requirements and needs of traffic data. To validate our hypothesis, experimental results have been presented and discussed with real traffic data captured over the city of Madrid (Spain), comprising up to 133 different datasets and a benchmark of 13 different forecasting methods, including shallow learning models, ensembles and alternative neural architectures. In light of the reported performance results and the statistical study of the discovered gaps, we conclude that Deep ESN should be widely embraced as a competitive method in future traffic forecasting studies, yielding further advantages such as the reduced complexity of its training procedure. 

Research efforts planned for the future will be invested towards evaluating the balance between complexity and performance of the latest Deep Learning methods applied to traffic forecasting. The ultimate goal is to provide empirical evidence on their acclaimed suitability for this specific application. Furthermore, a closer look will be taken at the explainability and understandability of RC methods, which are functional requirements of utmost importance to make forecasting models actionable in practice \cite{2002.02210,arrieta2020explainable}.

\section*{Acknowledgments}

The authors thank the Basque Government for its support through the EMAITEK and ELKARTEK funding programs, the Consolidated Research Group MATHMODE (IT1294-19), and the BIKAINTEK PhD support program (grant no. 48AFW22019-00002).

\bibliographystyle{IEEEtran}
\bibliography{biblio}
\end{document}